\documentclass{article}

\usepackage{arxiv}

\usepackage[utf8]{inputenc} 
\usepackage[T1]{fontenc}    
\usepackage{hyperref}       
\usepackage{url}            
\usepackage{booktabs}       
\usepackage{amsfonts}       
\usepackage{nicefrac}       
\usepackage{microtype}      
\usepackage{lipsum}
\usepackage{parskip}
\usepackage{graphicx}
\usepackage{float}
\graphicspath{{plots/}}
\usepackage{caption}
\usepackage{bm}
\usepackage[numbers]{natbib}

\title{Deep covariate-learning: optimising information extraction from terrain texture for geostatistical modelling applications}

\author{
  Charlie Kirkwood\\
  Department of Mathematics\\
  University of Exeter, UK\\
  \texttt{c.kirkwood@exeter.ac.uk} \\
}

\begin{document}
\maketitle

\begin{abstract}
Where data is available, it is desirable in geostatistical modelling to make use of additional covariates, for example terrain data, in order to improve prediction accuracy in the modelling task. While elevation itself may be important, additional explanatory power for any given problem can be sought (but not necessarily found) by filtering digital elevation models to extract higher-order derivatives such as slope angles, curvatures, and roughness. In essence, it would be beneficial to extract as much task-relevant information as possible from the elevation grid. However, given the complexities of the natural world, chance dictates that the use of `off-the-shelf' filters is unlikely to derive covariates that provide strong explanatory power to the target variable at hand, and any attempt to manually design informative covariates is likely to be a trial-and-error process --- not optimal. \par

In this paper we present a solution to this problem in the form of a deep learning approach to automatically deriving optimal task-specific terrain texture covariates from a standard SRTM 90m gridded digital elevation model (DEM). For our target variables we use point-sampled geochemical data from the British Geological Survey: concentrations of potassium, calcium and arsenic in stream sediments. We find that our deep learning approach produces covariates for geostatistical modelling that have surprisingly strong explanatory power on their own, with R\textsuperscript{2} values around 0.6 for all three elements (with arsenic on the log scale). These results are achieved without the neural network being provided with easting, northing, and absolute elevation as inputs, and purely reflect the capacity of our deep neural network to extract task-specific information from terrain texture alone. By visualising our deep-learned covariates as geographic maps, we can see that complex but general features of the surface environment and the subsurface are being captured. We hope that these results will contribute to further investigation into the capabilities of deep learning within geostatistical applications.
\end{abstract}

\keywords{Deep learning \and Geostatistics \and Terrain analysis \and Feature learning \and Convolutional neural networks \and Geology}

\section{Introduction}

Since it's inception in 1951 by mining engineer Danie Krige \cite{krige1951statistical}, the `kriging' method has largely defined the field of geostatistics. In the beginning, kriging was a purely spatial model, utilising only the spatial autocorrelation of the target variable in order to make new predictions. The underlying logic is perhaps best summed up by Tobler's third law of geography: that "everything is related to everything else, but near things are more related than distant things" \cite{tobler1970computer}. Kriging worked very well for its original purpose of interpolating gold grades in mines, where additional data was not available. In subsequent decades, as more data-rich problems began to be tackled, kriging evolved to include the ability to handle additional covariates in the model. There have been several somewhat muddled incarnations along the way (i.e. universal kriging \cite{matheron1969krigeage}, regression kriging \cite{odeh1995further}, kriging with external drift \cite{hudson1994mapping}) but, as has been well explained by Murray Lark \cite{lark2012towards}, in 1999 Michael Stein \cite{stein1999interpolation} brought mathematical clarity to the situation. Stein pointed out that all varieties of kriging can be considered as forms of the empirical best linear unbiased predictor (or BLUP) based on the linear mixed model:

\begin{equation}\label{kriging}
    \mathbf{Z = X\boldsymbol\tau + u + \boldsymbol\varepsilon},
\end{equation}

where $\mathbf{Z}$ is a random vector corresponding to the target variable at $n$ sites, $\mathbf{X}$ is an $n\times p$ design matrix, containing the values of any covariates, $\boldsymbol\tau$ are the corresponding fixed effects coefficients, $\mathbf{u}$ is a spatially correlated random variable (Gaussian process), and $\boldsymbol\varepsilon$ is an independently and identically distributed random variable. This formulation was significant for geostatistics as it enabled parameter estimation by maximum likelihood, with corresponding improvements over the previous method-of-moments approach \cite{lark2000estimating}. For our purposes, the formulation is significant because it allows us to separate the `regression on covariates' component, $\mathbf{X}\boldsymbol\tau$ --- on which we focus --- from the spatial, $u$, and noise, $\boldsymbol\varepsilon$, components of this definitive geostatistical model.

In general, obtaining measurements of the target variable for any geostatistical application is difficult. It could be said that this is the reason geostatistical models are required in the first place --- if we could easily observe our target variable at any point in space, we would have little need for statistical models. At the same time, the progression of technology has lead to a vast increase in data availability in general. In the geosciences, the rise of remote sensing means that multispectral satellite imagery is readily available for the entire globe, along with elevation data (which we make use in this study), and many countries have now conducted some form of airborne geophysical survey to provide gravity, magnetic, and radiometric measurements in continuous gridded format. Although these datasets tend not to measure our target variables directly, they may contain information that can contribute to the `regression on covariates' component of the typical statistical model, $\mathbf{X}\boldsymbol\tau$. But are we making the most of the information they provide? The typical geostatistical model, as formulated in \autoref{kriging}, is restricted to only being capable of capturing linear relationships between any provided covariates and the target variable. This means that the typical geostatistical approach to utilising remote sensing data has previously always had to involve manually post-processing gridded datasets in order to derive new covariates that we hope will be informative for the task at hand (i.e. that will display a linear relationship with the target variable).

In the case of terrain analysis, it is common to use a set of standard filters in order to obtain derivatives such as slope aspect, curvature and roughness. The Topographic Roughness Index (TRI) \cite{riley1999index} for example, has been used to identify landslides \cite{berti2013comparative}, model forest fire return levels \cite{stambaugh2008predicting} and map emerging bedrock in eroding landscapes \cite{milodowski2015topographic}, among other applications. But are we to believe that the TRI provides optimal explanatory power from the terrain to any of those tasks? The range of applications that have made use of `off-the-shelf' filters to derive covariates for their geostatistical models is huge. In fact, to the best of our knowledge, the study we present here is the first of its kind to demonstrate an approach for automatically deriving optimal task-specific covariates from gridded datasets for geostatistical modelling applications. The covariates we derive are optimal in that they have been engineered by our deep neural network to have maximal explanatory power with respect to the target variable, to which they relate linearly. This linearity ensures that our deep learned covariates are compatible for use within the fixed effects component of the typical geostatistical model, $\mathbf{X}\boldsymbol\tau$ in \autoref{kriging}.

In reality, if all the covariate information provided to the geostatistical model is being processed through our neural network, as it is in this study, then the neural network's output \textit{is} the fixed effect component of our geostatistical model. This is because of the 1:1 relationship (plus noise, $\boldsymbol\varepsilon$) between our neural network's output (the covariate, with values contained in $\mathbf{X}$) and the target variable, $\mathbf{Z}$. As a result, the value of $\boldsymbol\tau$, the fixed effect coefficient (singular in this case), would be one. We therefore replace our typical geostatistical model formulation with a `deep covariate-learning' geostatistical model formulation:

\begin{equation}\label{dcl}
    \mathbf{Z = D + u + \boldsymbol\varepsilon},
\end{equation}

where $\mathbf{D}$ is the output of our deep neural network. Because of the additive nature of these forumlations, we do not give much consideration to the spatial component, $u$, or the noise component, $\boldsymbol\varepsilon$, in the rest of this paper. They do not interact with $\mathbf{D}$, and so we focus on evaluating the explanatory power that our deep neural network output, $\mathbf{D}$, provides to our target variable, $\mathbf{Z}$, on a stand-alone basis.

\section{Method}

Our approach is inspired by the work of computer scientist Geoffrey Hinton and colleagues, who revolutionised the field of computer vision in 2012 by using deep learning to achieve a new state-of-the-art in image classification accuracy on the ImageNet Large Scale Visual Recognition Challenge (ILSVRC) \cite{krizhevsky2012imagenet}. Prior to their work, image classification problems had been solved by providing linear classification algorithms with sets of manually derived image features. Similarly to the way that `off-the-shelf' covariates are currently used in geostatistical modelling problems, it seemed unlikely that the manually derived image features were optimal for the task at hand, but a viable alternative had yet to be proven. Deep learning changed everything by replacing the existing setup with end-to-end learning: in deep learning the classification algorithm is also the feature learner --- feed raw images in, and get answers out. In 2012, the answers that Krizhevsky, Sutskever, and Hinton got out --- correct labels for images --- were the best that had ever been achieved \cite{krizhevsky2012imagenet}, and lead to the ubiquituos use of deep learning in computer vision applications. We hope that the parallels between manually creating features for image classification and manually deriving covariates for geostatistical applications are apparent.

In our case, we want to learn features from terrain texture in a similar way, so that we can go beyond using `off-the-shelf' terrain derivatives as covariates, and extract more explanatory power from the landscape for any specific task. To do this we use our own deep neural network, constructed from similar building blocks as used by Hinton and colleagues in 2012. The critical difference is that in our case we want to learn to do image \textit{regression} rather that classification, because our target variables (element concentrations from geochemical survey data) are continuous. In practice this simply means giving our network a single linear output rather than using a multinomial logistic output. An additional consideration for us has been the importance of retaining spatial context amongst the terrain texture. Translation invariance is an important feature of the deep neural networks used in image classification: it shouldn't matter where in the image the cat is, it's still an image of a cat. For our purposes however, it seems likely that the positions of terrain features relative to our prediction point and to each other matters greatly. In geochemistry, concentrations of immobile elements can be expected to be associated in situ with certain bedrock types, while mobile elements may show spatial relationships with distance to faults and other fluid conduits along which they might be mobilised.

It should be noted at this point that deep learning has been applied to problems within the realms of remote sensing and geostatistics before, and the novelty of our study does not lie in deep learning itself but in how we use it. For some background, deep learning has seen significant use in remote sensing applications in the latter half of the last decade, applied to tasks of object detection, scene classification, image fusion, image registration, land-use classification, semantic segmentation and more \cite[e.g.][]{han2014object,zou2015deep,zhang2016deep,zhu2017deep,ma2019deep}. Meanwhile, general machine learning approaches have been applied to the spatial interpolation of environmental variables \cite[e.g.][]{li2011application}, which had traditionally been considered the preserve of geostatistics. Complex mapping tasks, such as that of landslide susceptibility \cite{pourghasemi2018prediction}, or mineral prospectivity analysis \cite{rodriguez2015machine} have always relied on being provided with good covariates in order to achieve good results. It is exactly to these complex mapping tasks, where a ground-measured target variable is modelled with the support of remotely-sensed auxiliary variables, that our deep covariate-learning approach appeals. Where previously covariates have had to be derived manually from remotely-sensed auxiliary variable grids, deep covariate-learning allows this covariate-derivation process to happen automatically and optimally. The unique contribution of this study is therefore that, to the best of our knowledge, it is the first to show how the feature-learning ability of deep learning can be used within the framework of the well-established BLUP geostatistical model (\autoref{kriging}), thus providing new capabilities for use in geostatistical applications.

We believe that the interface between deep learning and geostatistics is an under-explored area in general, but would like to highlight some contributions that have preceded us in this space: \citet{wadoux2019using, wadoux2019multi} and \citet{padarian2019using} have demonstrated how deep learning can be used in the context of digital soil mapping, and their work has utilised the feature-learning ability of deep learning (although manually derived covariate grids are also included). However, where as we present deep learning as a way to learn optimal covariates for use in typical geostatistical models --- as deep covariate-learning --- these previous studies have used deep learning to replace the entire geostatistical model. As we discuss later on, there may well be benefits to such end-to-end approaches. Nevertheless, we believe our deep covariate-learning approach provides a unique contribution to this under-explored research area in that it lays bare the ability of deep learning to extract information from remotely-sensed auxiliary variable grids even in the absence of explicit spatial location information. We hope this work will contribute to continued investigation into how to combine the best of both worlds (geostatistics and deep learning) in order to advance our capabilities in modelling and mapping complex environmental phenomena.

\subsection{Data setup}

\begin{figure}[!htb]
    \centering
    \includegraphics[width=0.495\textwidth]{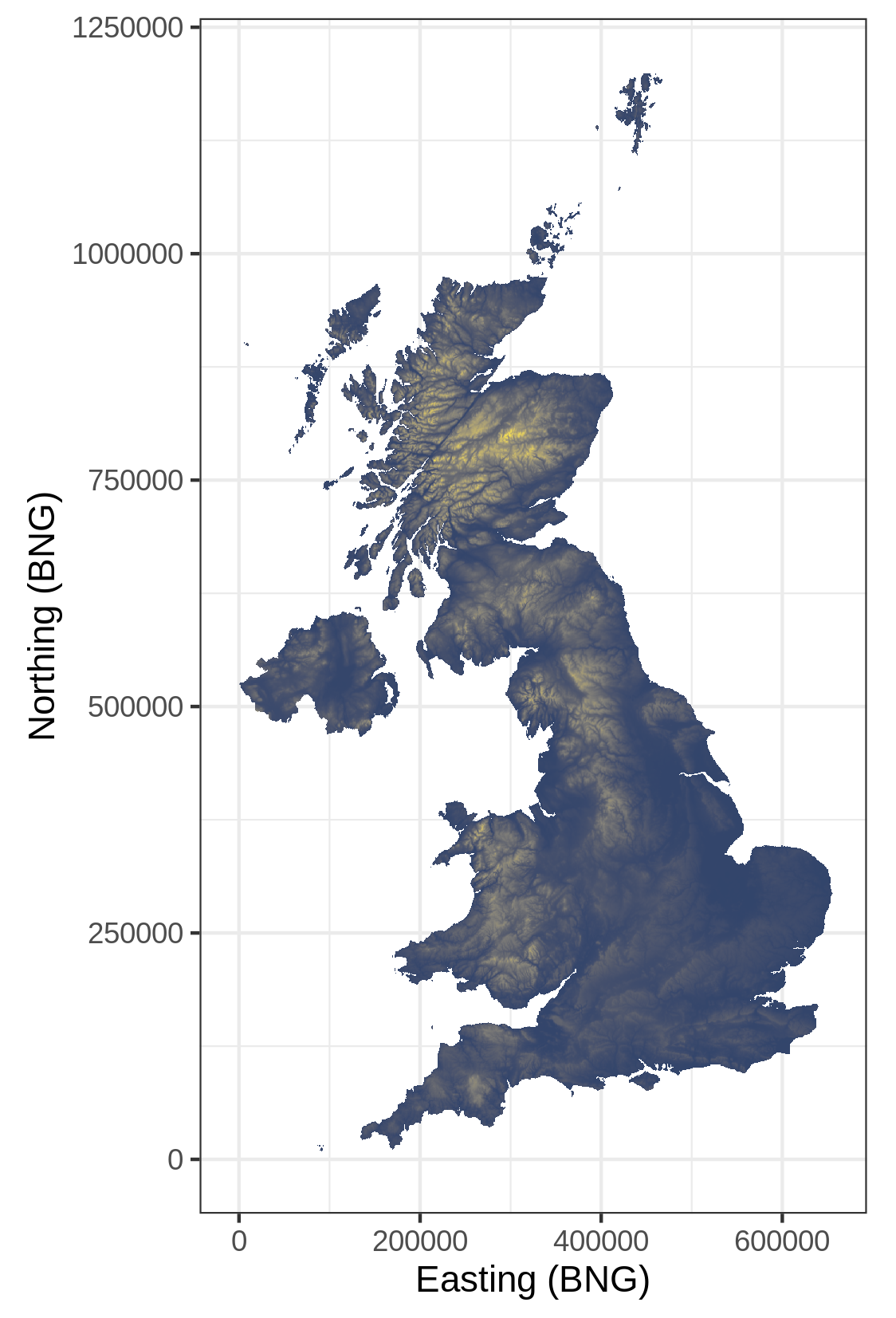}
    \includegraphics[width=0.495\textwidth]{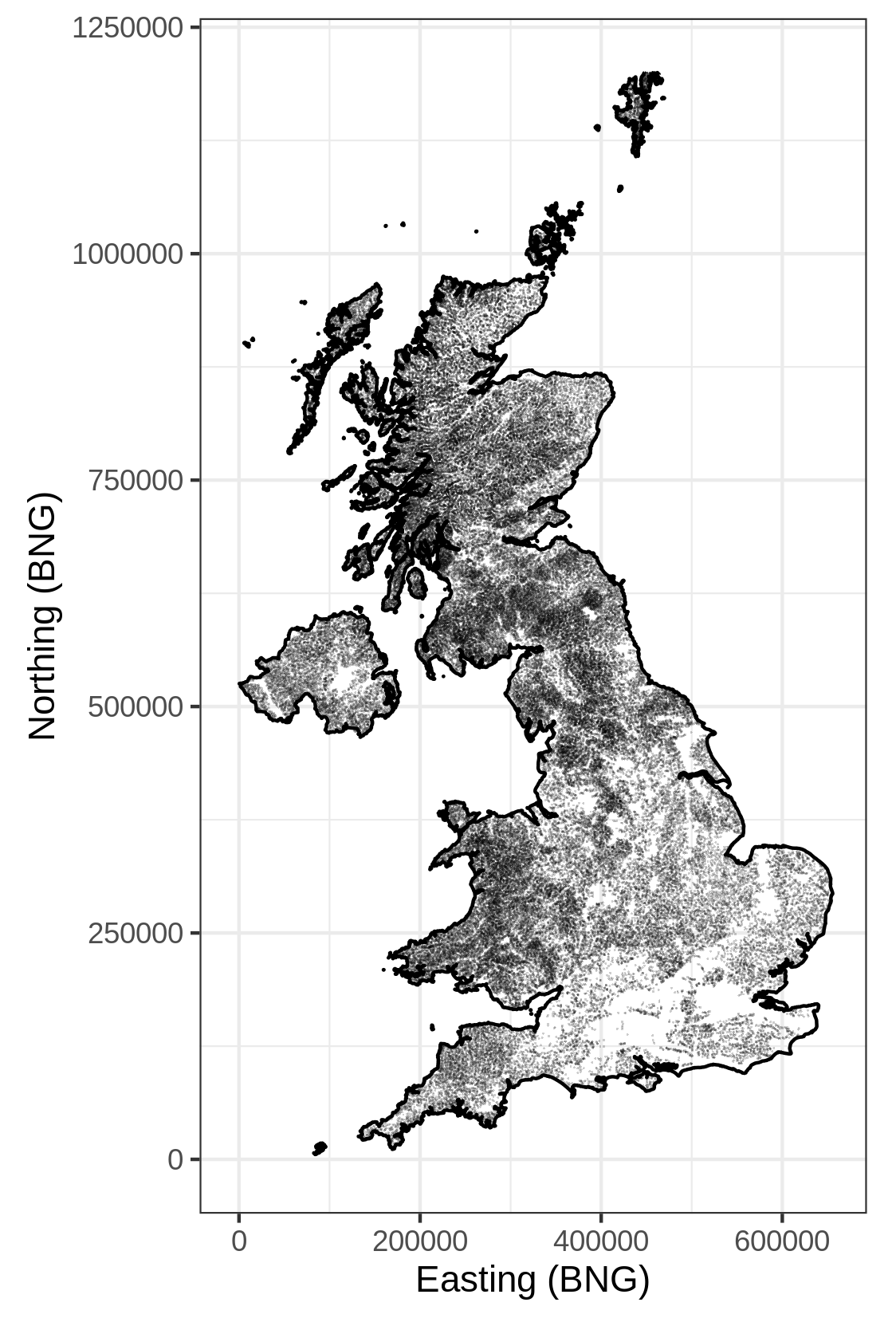}
    \caption{\textbf{Left -} SRTM 90m elevation data for the UK. \textbf{Right -} G-BASE stream sediment geochemistry sample collection sites (110 794 sites in total).}
    \label{fig:data}
\end{figure}

For this study, we make use of two datasets: 1) NASA's SRTM 90m gridded global elevation data \cite{van2001shuttle}, from which we use deep learning to derive optimal covariates in order to map 2) element concentrations from the British Geological Survey's G-BASE stream sediment sampling program \cite{johnson2005g}. Both can be seen in Fig.\ref{fig:data}. The geochemical dataset contains element concentrations from 110 794 sample sites from across the UK, though the number of observations used in this study varies by element as sites containing NA values are excluded. Any element concentrations reported below the accepted lower limit of detection were set to half the lower limit of detection as in previous studies using the geochemical dataset \cite{kirkwood2016stream}. For readers whose focus is on geochemical mapping and prospectivity analysis, we would recommend using log-ratio transformations on the geochemical data to avoid issues with compositional closure \cite{pawlowsky2015modeling}. However, in this study our focus lies in learning terrain textural features, and so we simply use our element concentrations in their raw form, with the exception that we log transform arsenic in order to improve stability of the gradient descent process by which the neural network is trained, and to make for more eye-friendly visualisations.

\begin{figure}[!htb]
    \centering
    \includegraphics[width=0.31\textwidth]{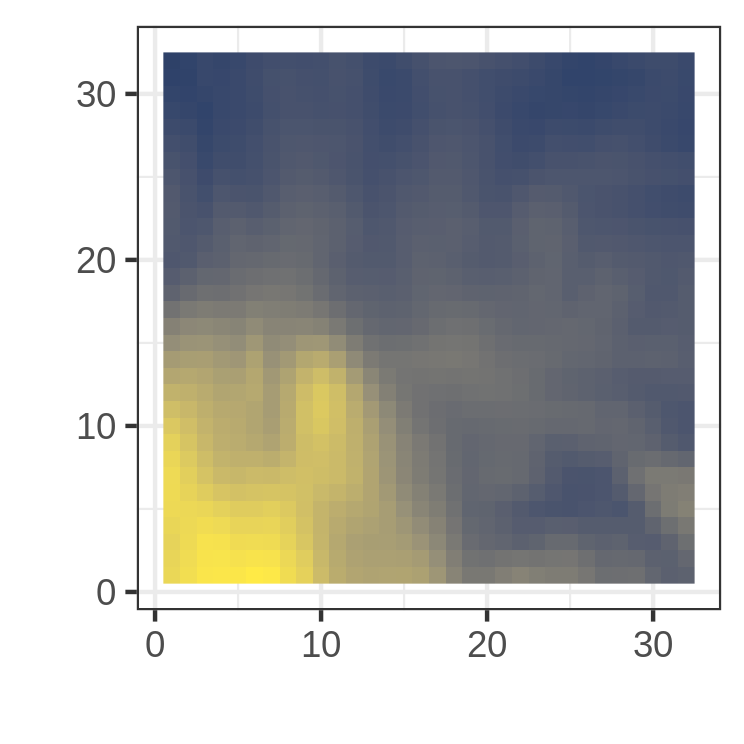}
    \includegraphics[width=0.31\textwidth]{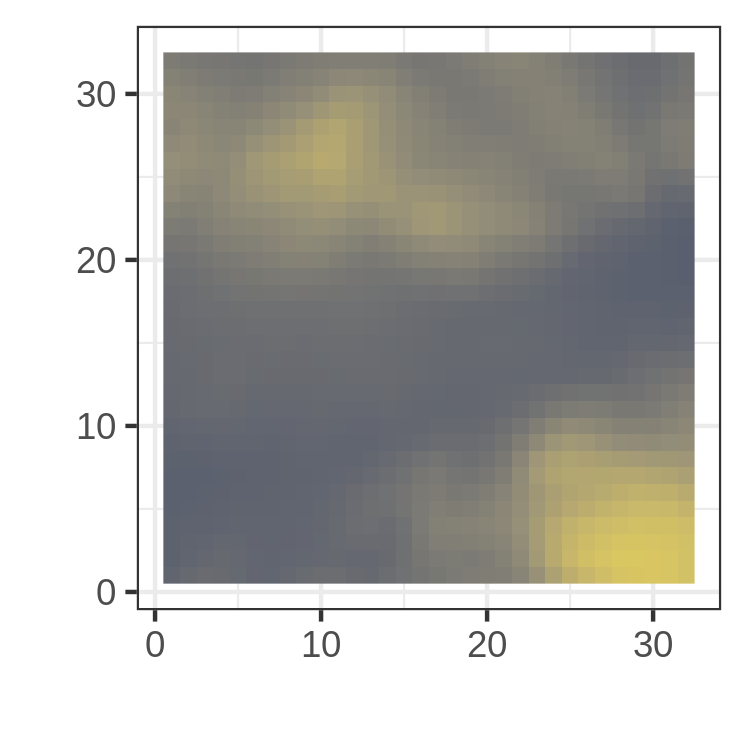}
    \includegraphics[width=0.31\textwidth]{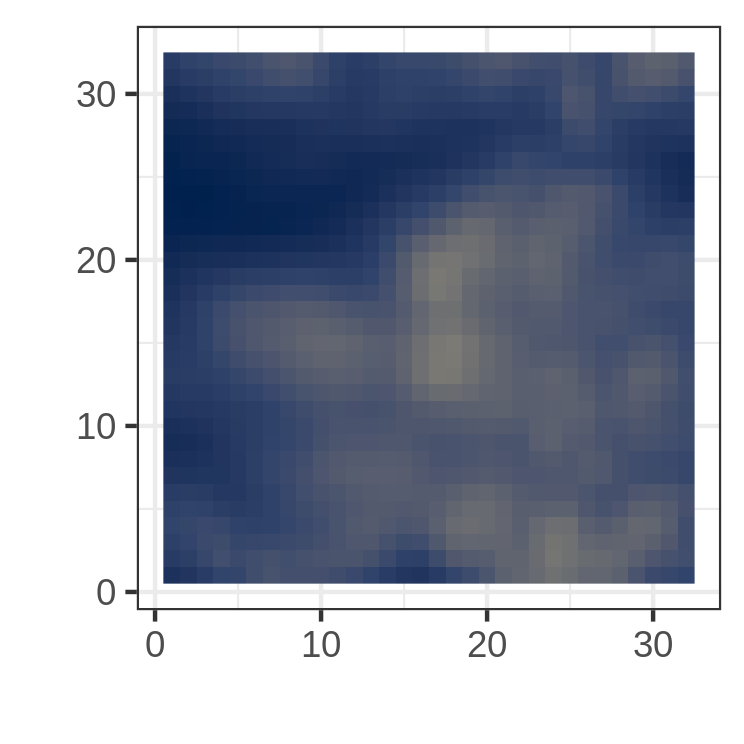}
    \caption{Three examples of 32x32 cell terrain input images as provided to our neural network. The colour scale is linear with cell elevation and is shared across all three images - more extreme shading variations therefore represent more extreme terrain. However, the absolute elevation of each terrain image has been normalised out, so that the central point is always at zero. This means that the neural network cannot use absolute height to `cheat' - it must learn features purely from the terrain texture.}
    \label{fig:imgs}
\end{figure}

While our target variables are simple element concentrations, our inputs are square images of SRTM terrain data. For our training dataset, these images each consist of a 32x32 cell window of terrain centred around their respective geochemical sample site (Fig. \ref{fig:imgs}). We use a cell size of 500m, which gives a real-world window size of 16x16km square. The size and resolution of these images can be thought of as a tunable hyper-parameter to the neural network, but in reality we chose them by our own visual judgement, believing that they should provide a reasonable amount of information without exceeding our compute capacity (an Nvidia Titan X Pascal GPU - with thanks to Nvidia's grant scheme). As is standard practice in neural network training, we normalised our input data values. This is typically done to each input variable by subtracting the mean and dividing by the standard deviation, in order to achieve a mean of zero and a standard deviation of one. In our case we set the centre of each image to zero and divided all elevation values by the standard deviation of the UKs elevation grid. By setting the centre of each image to zero, we remove elevation as an explicit variable to the neural network. We also don't provide easting and northing to the neural network - terrain texture is all it has to make use of.

For each element, our dataset therefore consists of an element concentration vector of length $n$ and a corresponding multidimensional array of dimensions $n\times32\times32\times1$ that contains the images to be input to the neural network. It is worth mentioning that while our images only contain a single channel (terrain) there is no reason why our approach cannot be extended to multiple channels if other continuous covariates are available (such as from other airborne and satellite surveys).

\subsection{A deep neural network for terrain filtering}

\begin{figure}[!htb]
    \centering
    \includegraphics[width=\textwidth]{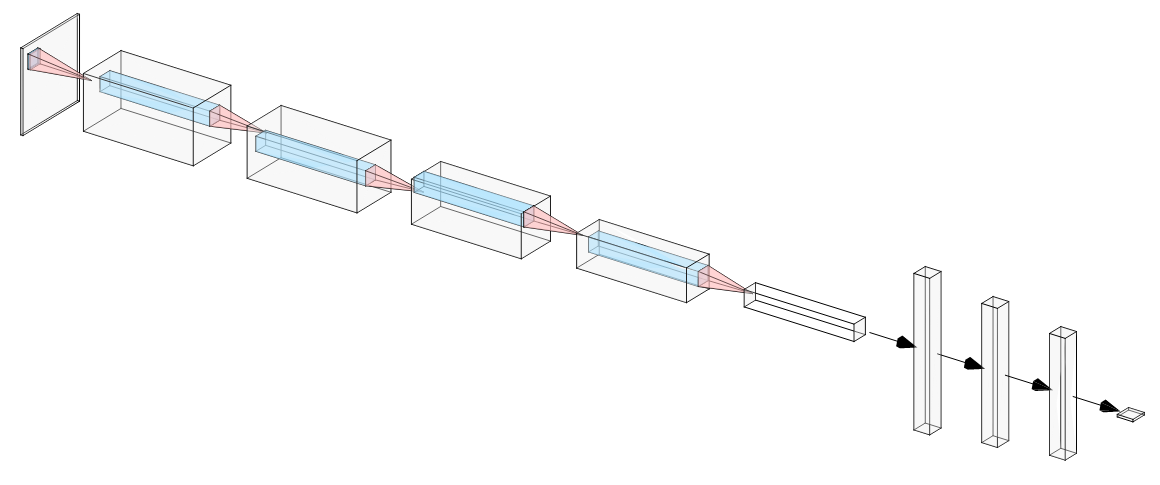}
    \caption{The architecture of our deep neural network. Input terrain images of size 32x32x1 (left hand side) are filtered through 5 128-channel convolutional layers and a single average pooling step to represent each image as a 4x4x128 spatial tensor. This then flattened into a vector of length 512 before being passed through two more fully connected hidden layers (256 and 128 nodes) prior to the final output - a single linear output (right hand side). The network uses dropout throughout, and a small amount of gaussian noise is added before each convolutional layer, to minimise overfit.}
    \label{fig:DNN}
\end{figure}

We implement our deep neural network using the Keras interface to Tensorflow, via the R language for statistical computing. We refer readers to our code for full details. The architecture of the network we present here represents the best performance we were able to achieve through fairly extensive trial and error. For future versions we may utilise automated procedures for architecture design and hyper-parameter tuning, but it was an enjoyable experience to gain intuition into effective neural network designs for extracting information from the terrain. The design we settled upon (Fig. \ref{fig:DNN}) consists of a series of stacked convolutional layers topped off with an average pooling layer which feeds into a fully-connected multilayer perceptron-type architecture which provides the final output (predictions of element concentrations). In total our network has just over 600 000 trainable parameters, and our objective function is to minimise mean-squared-error (MSE) in relation to the target variable. In order to prevent overfitting, we use dropout at every level in the network, and inject a small amount of gaussian noise ahead of each convolutional layer to further aid generalisation. Despite our efforts, it seems almost certain that the design we present here is \textit{not} truly optimal (indeed, the optimal network design would be different for any given set of data) but it performs well in our application. The field of deep learning is very fast moving, and in this study we aim to share the general approach of using deep learning to derive task-specific covariates from terrain texture, rather than promote any particular network architecture.

\begin{figure}[!htb]
    \centering
    \includegraphics[width=\textwidth]{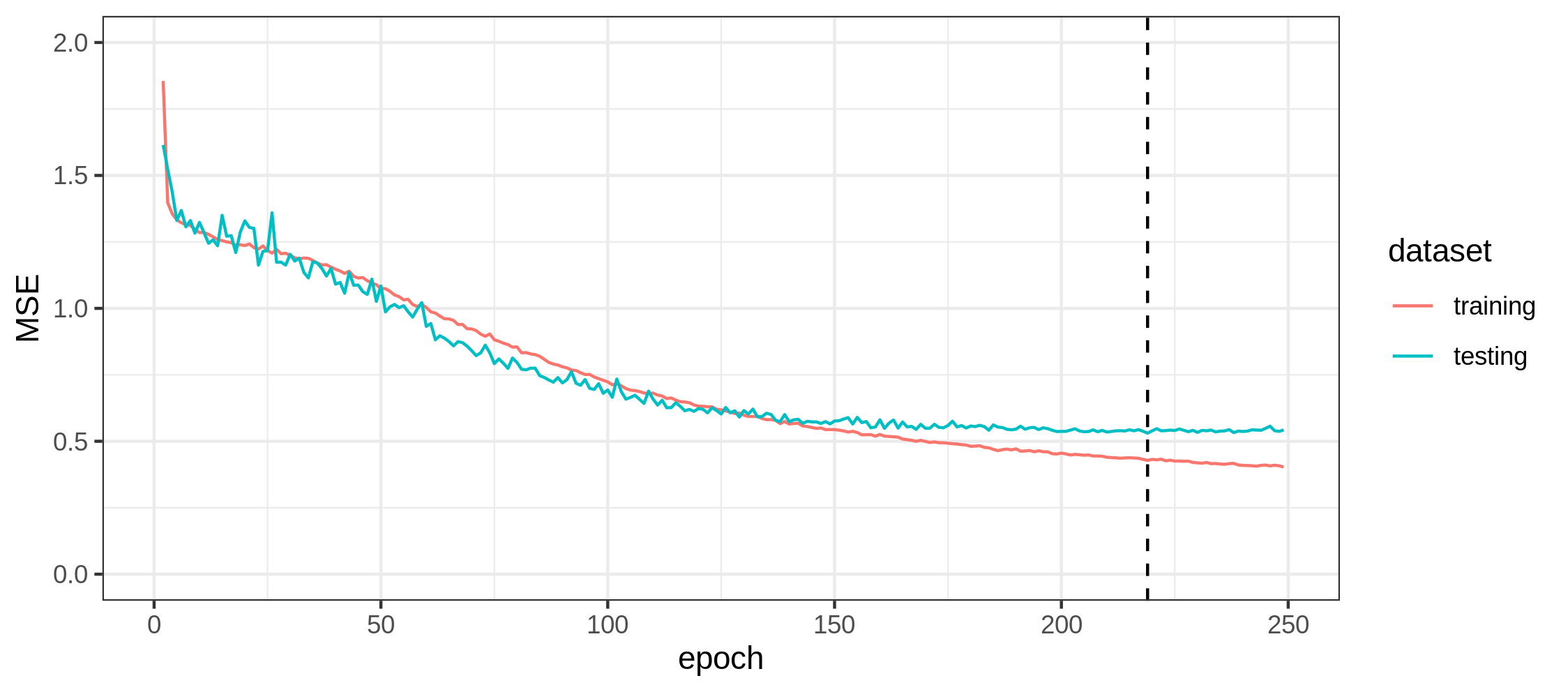}
    \caption{The training history of our neural network trained to predict log(arsenic) in stream sediments from terrain texture. The vertical dashed line marks the best epoch, for which the mean-square-error (MSE) on held out test data is lowest. The weights at this best epoch are the ones that are kept for subsequent use.}
    \label{fig:training}
\end{figure}

To train the neural network, for each element we split our dataset into 10 folds at random, and trained using 9 of them, while monitoring the mean-squared-error (MSE) of the neural network's predictions on the 10\textsuperscript{th} fold to ensure that we did not overfit. We trained the neural network using the ADAM optimiser, and a batch size of 4096 observations. We ran training for up to 300 epochs (Fig. \ref{fig:training}), but early stopping tended to find the best fit around 200 epochs (before the MSE began to increase again on held out test data as the network began to overfit, but this was very gradual thanks to our regularisation measures). On our NVIDIA Titan X Pascal GPU each training run (one for each element) took about 10 minutes.

Once our deep neural network has been trained to predict the concentration of an element from the terrain (though this could equally be any other target variable), its output \textit{is} the optimal terrain texture covariate that we wanted to learn. As we saw in \autoref{dcl}, if no other covariates are supplied to the geostatistical model, as is the case in this study, then the deep neural network output in fact becomes the entire fixed effect component of our `deep covariate-learning` statistical model. The neural network knows nothing of the spatial location at which a prediction is to be made (no easting, northing, or absolute elevation were provided) and can only extract information contained within the texture of surrounding terrain. It stands in for the role of the `off-the-shelf` covariates in the typical geostatistical model, with the aim to provide as much explanatory power as possible independently from the spatial component of the problem. The difference is that by using deep learning we are able to optimise this process of extracting information from the terrain --- the neural network learns to derive the best terrain texture covariates that it can for the task at hand --- providing as much explanatory power as possible with respect to the target variable.

\section{Results}

\begin{figure}[!htb]
    \centering
    \includegraphics[width=0.31\textwidth]{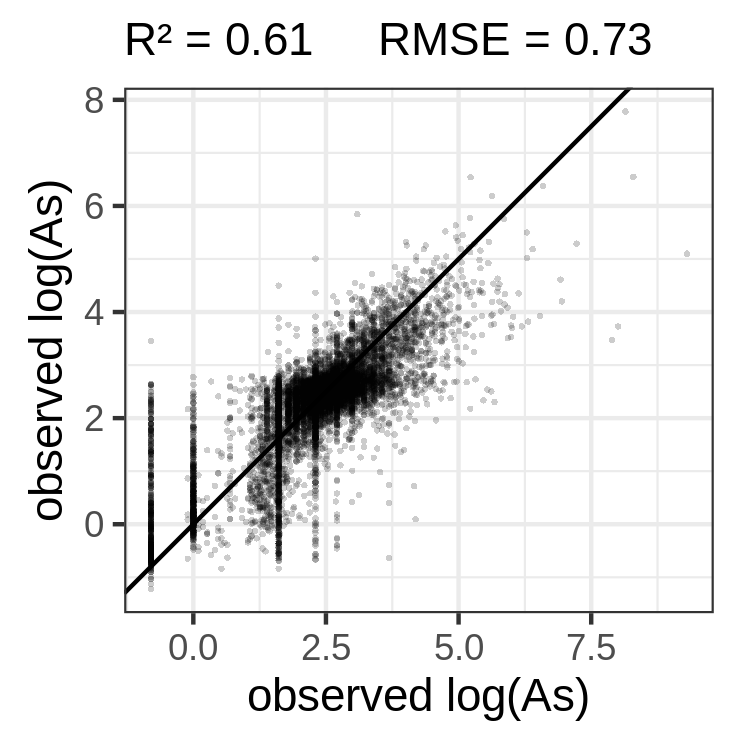}
    \includegraphics[width=0.31\textwidth]{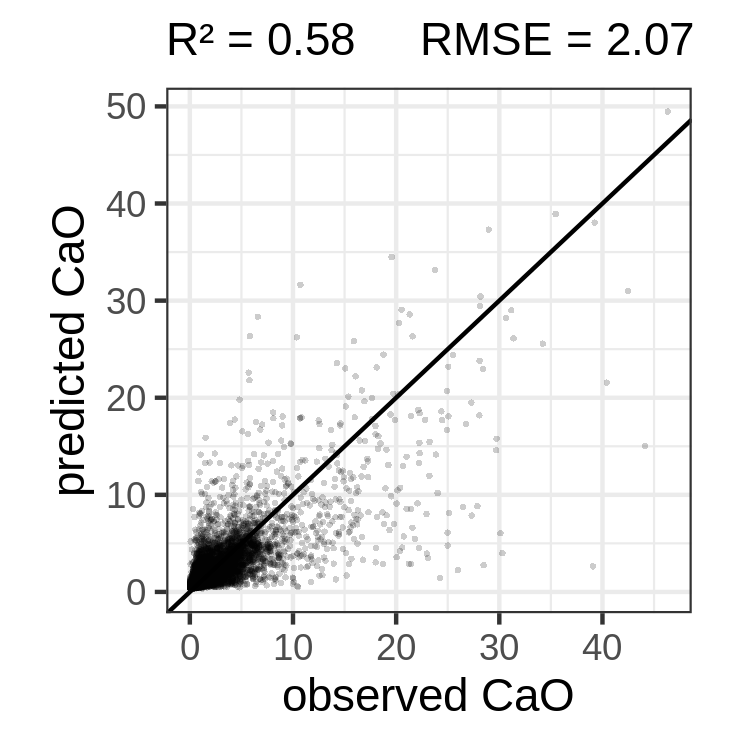}
    \includegraphics[width=0.31\textwidth]{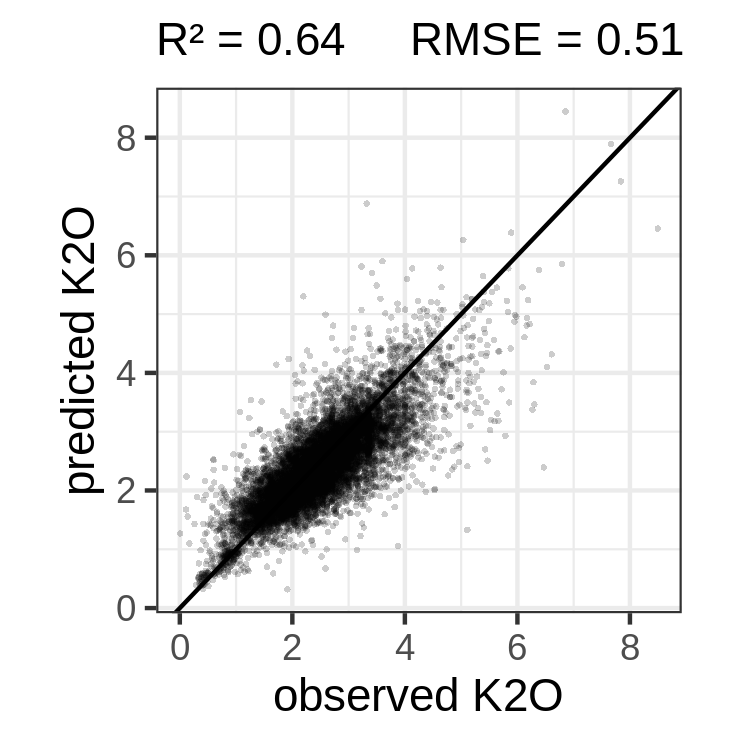}
    \caption{Plots evaluating the predictive performance, or explanatory power, of the output of our deep neural network trained to optimally extract terrain texture information in order to predict each of our three target variables: arsenic (As), calcium (CaO), and potassium (K2O) concentrations in UK stream sediments. These evaluations are made on held out test data that was not seen by the network during training.}
    \label{fig:eval}
\end{figure}

We can evaluate the explanatory power of our neural network's output by comparing its predictions to the true observed values in held-out test data (the 10\textsuperscript{th} fold - not used in training, \autoref{fig:eval}). By doing so we find that our deep learning approach is able to explain a very significant proportion of the variance in our target variables. It explains 61\% of the variance in log(arsenic) (As), 58\% of the variance in calcium (oxide, CaO), and 64\% of the variance in potassium (oxide, K2O). These are impressively high degrees of explanatory power to achieve by harnessing the information contained within terrain texture alone. Geologists have long understood that underlying geology is reflected in the terrain, but the complexity of this relationship --- requiring caveats, conditions, and qualifiers at every turn --- has never lent itself to being formalised. It is somewhat remarkable then, that deep learning has been able to capture this relationship so successfully, and in less than ideal circumstances given that our neural network has never been told where it is - all it gets to see is the 16x16km window of terrain - and always centred at zero elevation.

To get a feel for the complexity of terrain features that the neural network has been able to learn (each in relation to the concentrations of chemical elements in stream sediments) we can generate maps of its output. We do this by making predictions from the neural network on a regular grid. For each prediction, the corresponding 16x16km terrain window is first extracted from the underlying SRTM elevation data (and elevation normalised, as explained in methods), which are then provided to the neural network so that it can make predictions for the new locations. Even though the neural network has never seen these new windows of terrain before, we take its performance on the held-out test set (Fig. \ref{fig:eval}), which it had also never seen, as evidence of its explanatory ability on previously unseen data. The maps we produce in this manner are essentially SRTM elevation grids run through a complex non-linear filter (machine-learned from the bottom up, not designed from the top down) which maximises explanation of the target variable.

\begin{figure}[!htb]
    \centering
    \includegraphics[width=\textwidth]{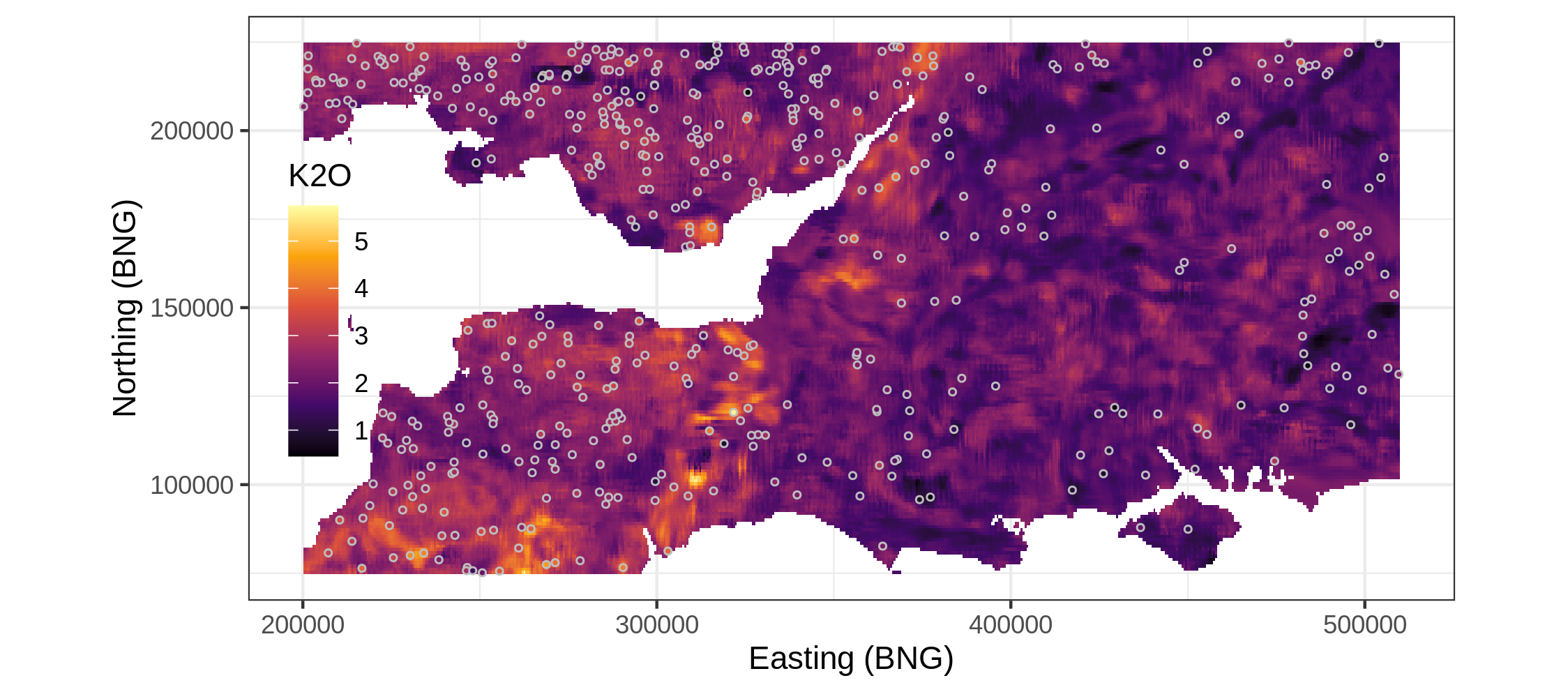}
    \includegraphics[width=\textwidth]{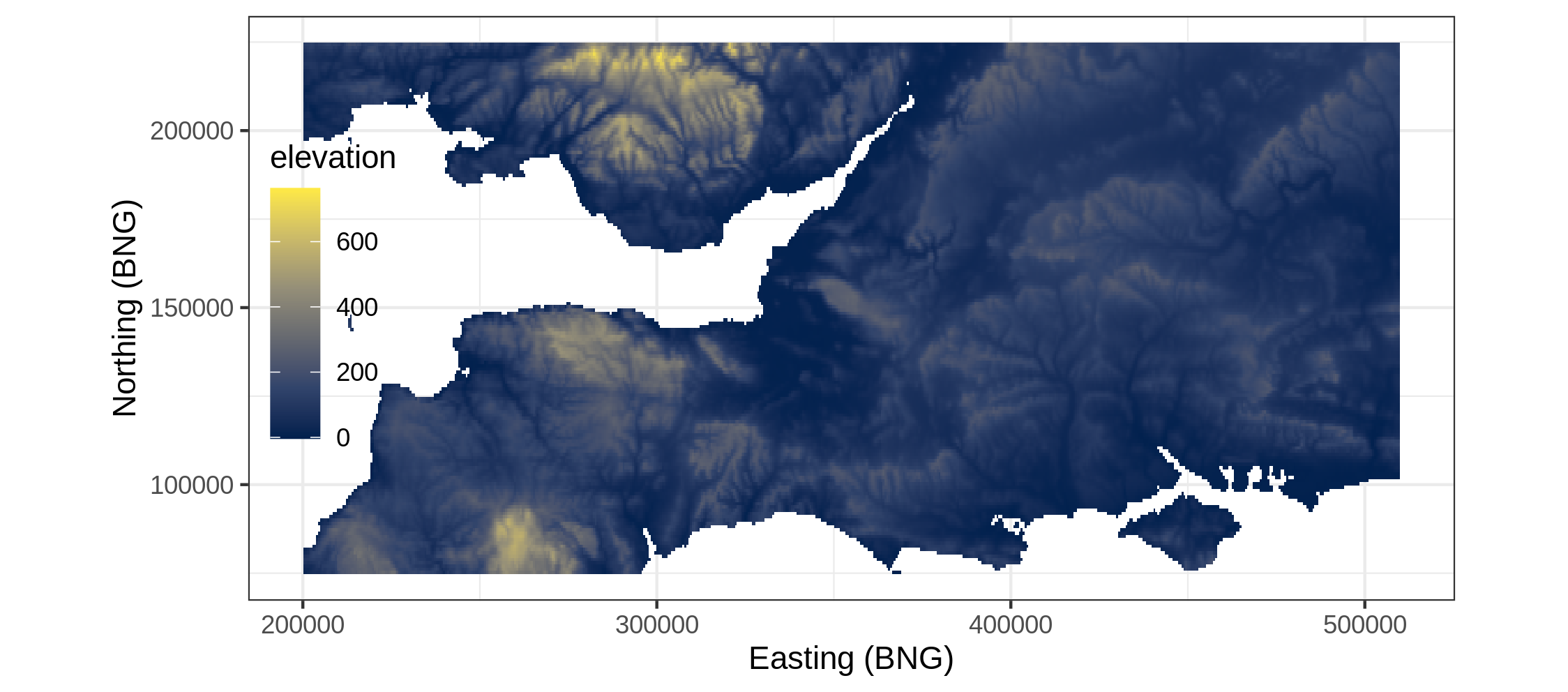}
    \caption{\textbf{Top -} Map of potassium concentrations as predicted exclusively from terrain texture using our deep neural network. 500 random geochemical sample sites are overlain. These ground-truth point values share the same colour scale as the raster map. The lack of deviation between the ground-truth and the prediction (also seen in Fig. \ref{fig:eval}) supports the conclusion that the detail in the map is `real' and not a product of over-fitting. However, at these scales some checker-board aliasing artefacts are apparent, which we would hope to remove with subsequent refinement of our neural network architecture. \textbf{Bottom -} The corresponding SRTM terrain from which the above map is derived via deep learning.}
    \label{fig:local}
\end{figure}

\begin{figure}[htbp]
    \centering
    \includegraphics[width=\textwidth]{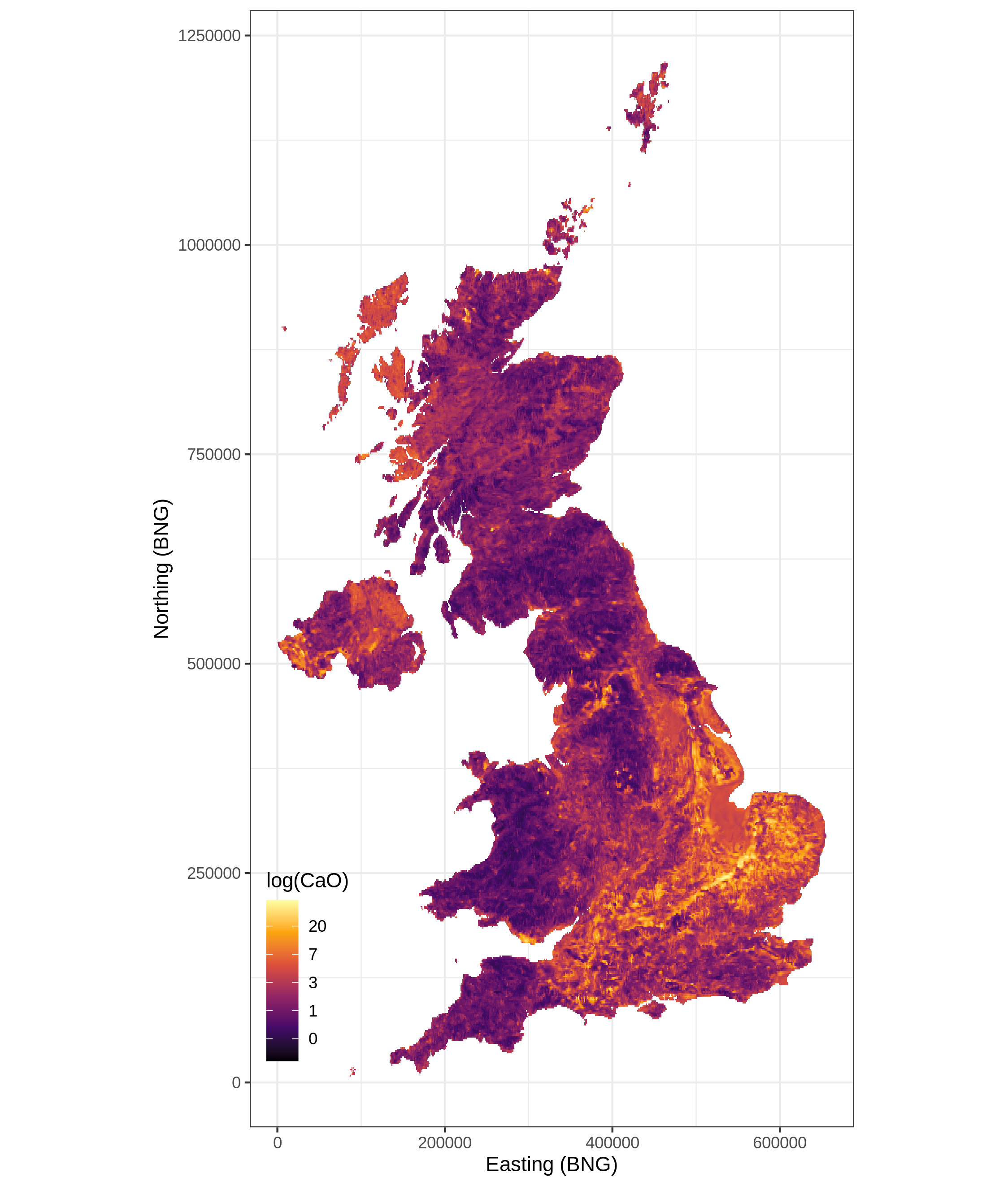}
    \caption{Our optimal terrain texture covariate for the prediction of stream sediment calcium. The map was produced by running the UK's SRTM elevation grid through our deep-learned terrain texture filter, optimised for explanatory power with respect to calcium concentrations. This map accounts for 58\% of the variance in calcium concentrations through terrain texture alone. Subsequent geostatistical modelling can be used to improve prediction further, by taking account of spatial information (easting, northing, elevation) and perhaps other non-terrain based covariates too.}
    \label{fig:UKfull}
\end{figure}

\begin{figure}[!htb]
    \centering
    \includegraphics[width=0.495\textwidth]{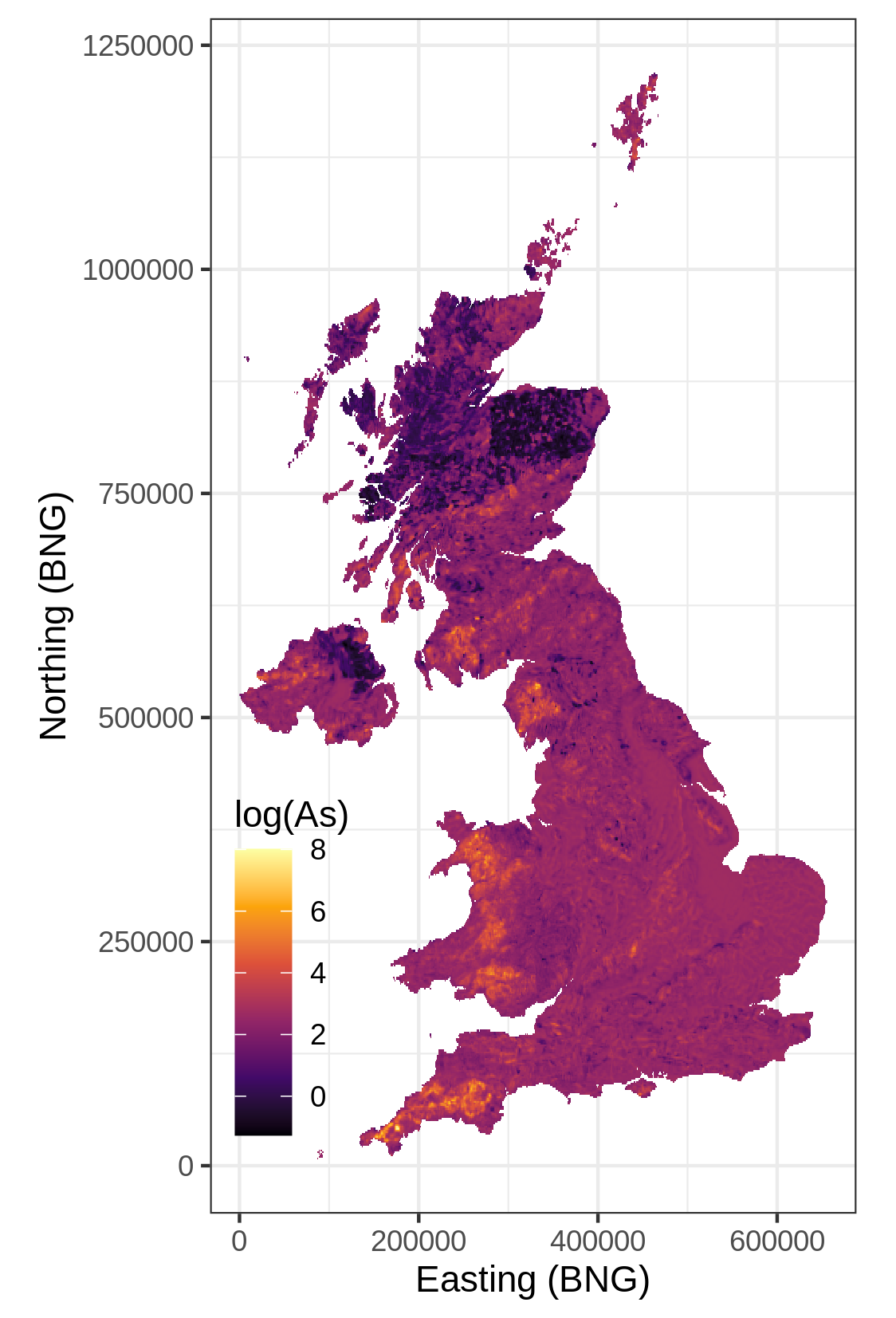}
    \includegraphics[width=0.495\textwidth]{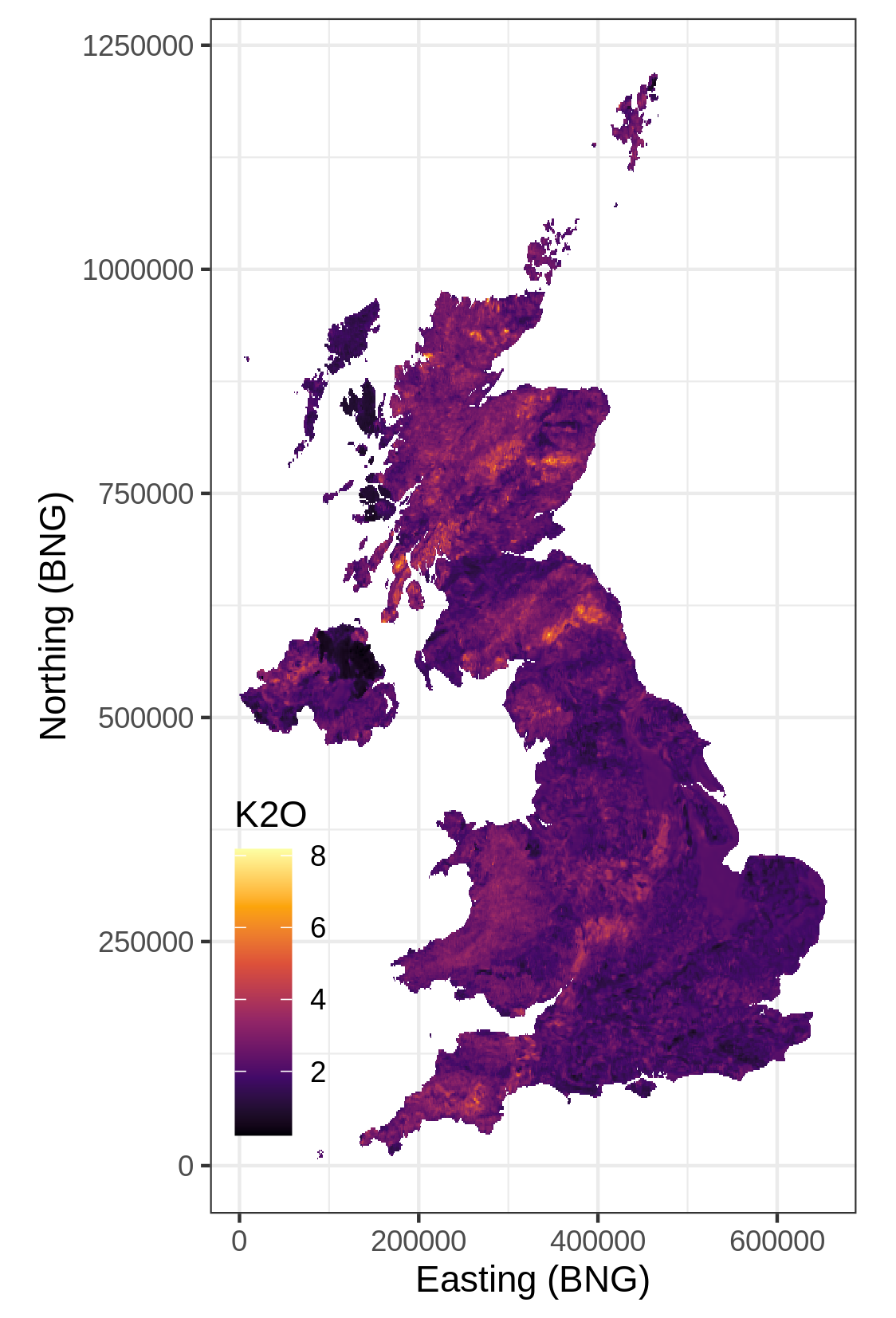}
    \caption{\textbf{Left -} Our optimal terrain texture covariate for the prediction of stream sediment arsenic concentrations. \textbf{Right -} Our optimal terrain texture covariate for the prediction of stream sediment potassium concentrations. In both cases, geological features are clearly apparent, and it is fascinating to see these being revealed through geochemically-optimal filtering of terrain texture alone. We recommend the British Geological Survey's iGeology mobile app (https://www.bgs.ac.uk/igeology/) to readers who wish to learn more about the features that these maps reveal.}
    \label{fig:UKhalf}
\end{figure}

These deep-learned covariate maps (Fig. \ref{fig:local}, \ref{fig:UKfull}, \ref{fig:UKhalf}) reveal a great deal of geological information, but in fact their task of explaining stream sediment geochemistry is more complex than explaining geology alone: Not only are stream sediments subject to the influence of surface processes as well as geological ones, but they also consist of mixtures of material accumulated from their upstream catchment area rather than representing any single point. This areal property makes their prediction difficult without accounting for upstream catchments during any modelling process \cite{kim2017geostatistical}. Visually (and it will be very interesting to investigate further), our deep-learned covariate maps do appear to have captured some flow-like effects. For example, in Fig. \ref{fig:UKfull} we can see patterns that appear to show the `washing out' of elevated calcium concentrations from the chalk scarp that brightly trends north-east into East Anglia (the most eastern lobe of the UK). While these maps --- which harness \textit{only} the information contained within terrain texture --- fall short of explaining all of the variance in our target variables (which would never be expected) it is somewhat surprising that they are able to explain so much of it (with R\textsuperscript{2} values around 0.6 for all three elements), and a very encouraging result for the use of deep learning in geostatistical applications.

\section{Discussion}

Our results have shown that deep neural networks are capable of extracting a great deal of geochemically-explanatory information from terrain texture alone, and it seems likely that similar success could be had by applying our methodology to other target variables and input grids. Adding additional channels to our terrain input images where available, such as for gravity, magnetics, and radiometrics data would likely further improve the neural network's ability to explain geochemistry and perhaps other target variables too. It will be very interesting to explore how widely applicable this deep covariate-learning approach is in future research. Could deep learning revolutionise geostatistics the same way it revolutionised computer vision in 2012? Some encouraging evidence comes from the fact that previous investigations of machine learning for geochemical mapping (but using `off-the-shelf' covariates) \cite{kirkwood2016dropout,rodriguez2015machine,kirkwood2016machine,zuo2017machine,kirkwood2017unmixing} have generally found terrain data to be among the least informative when compared to data from geophysical surveys \cite{kirkwood2016machine}. It will be interesting to see what deep covariate-learning can achieve when applied to these innately more geochemically-informative datasets.

In a sense we do injustice to deep learning in this study by not treating our application (modelling element concentrations) as an end-to-end problem: we have only tasked the neural network with the restricted function of learning to derive optimal terrain texture covariates for use in geostatistical models (i.e. either to contribute to $\mathbf{X}$ in \autoref{kriging}, or to take the place of $\mathbf{X}$ as $\mathbf{D}$ in \autoref{dcl}), rather than tasking the neural network with replacing the geostatistical model entirely. To do so would require that the neural network also handles the spatial component of the problem. This could be achieved most conveniently by simply providing the neural network with easting, northing, and absolute elevation as additional input variables. The neural network would effectively then replace both $X\boldsymbol\tau$ and $u$ in our model formulation. It is actually likely that doing so would result in improved prediction accuracy over the geostatistical model by virtue of the fact that the neural network would be free to learn the interactions between terrain features and spatial location. Conversely, the additive nature of the statistical model formulation (\autoref{kriging}, \autoref{dcl}) prevents interaction between the spatial component, $u$, and the fixed effects `regression-on-covariates' component, $\mathbf{X}\boldsymbol\tau$. This is perhaps a detrimental over-simplification, particularly for large and heterogeneous study areas, although it is well established practice nevertheless.

The reasons we have not gone all the way to providing an end-to-end `complete solution' neural network in this study are two-fold. Firstly, at this stage we find more scientific interest in investigating the ability of deep-learning to derive optimal covariates for geostatistical modelling, given that the use of covariates in geostatistical models (i.e. \autoref{kriging}) is such well-established practice. As it is, our `deep covariate-learning' geostatistical model formulation (\autoref{dcl}) seems like a reasonable middle ground from which to investigate the opportunities of deep learning within geostatistics without having to leave the established geostatistical modelling framework behind. This brings us to the second reason for not presenting an end-to-end solution here: While an end-to-end approach would play into the defining strength of deep learning --- its unique ability to learn features from unstructured data in order to optimise an objective --- it would also reveal what is currently deep learning's main weakness: uncertainty quantification. Uncertainty quantification in deep learning is a rapidly developing sub-field and promising breakthroughs have been made \cite{gal2015dropout,kendall2017uncertainties,farquhar2019radial}, but at the time of writing, it is likely that potential end users of our approach would prefer to use deep learning to derive optimal covariates for use within well-established geostatistical model formulations (e.g \autoref{kriging}), hence the title and angle of this paper.

Despite the restricted capacity within which we apply deep learning in this study (i.e. to learn optimal covariates for geostatistical modelling, rather than using deep learning as an end-to-end solution in itself), the implications of our results are very significant. Let's take mineral exploration for example, although similar situations are likely to occur in other applications: The original geostatistical approaches (still often used), which rely purely on the spatial auto-correlation of the target variable are almost destined to perform poorly in the search for new mineral deposits. This is because they can only interpolate between observations in the geographic space. In such cases, if we have not been fortunate enough to `hit' a mineral deposit with one of our samples, then the deposit can easily remain unseen between sampling locations. Adding a `regression-on-covariates' fixed effects component to the geostatistical model (e.g. \autoref{kriging}) alleviates this pathology to an extent, but only in as much as the available covariates can provide a good explanation of the target variable. Using deep learning to derive optimal covariates is therefore a step-change in the geostatistical modelling approach, as it allows us to objectively optimise the explanatory power we can obtain from gridded auxiliary datasets for any geostatistical modelling task. In doing so, we are able to explain the distribution of our target variable in terms of the deeper relationships between the target variable and terrain properties. The results we have obtained in this study demonstrate that these deep relationships do generalise spatially. This is evidenced by the fact that our neural network achieves the explanatory power that it does without ever being provided with easting, northing, and absolute elevation by which to infer its spatial position. The patterns that it learns to recognise between terrain texture and geochemistry therefore have to be applicable throughout the study area. This feature of deep covariate-learning therefore makes it an exciting new tool for identifying undiscovered mineral deposits, assuming that some examples of known mineral deposits are included within the training data. Based on our results, we would not be surprised to see deep learning become a key technology for discovering the mineral deposits of the future, each always harder to find than the last.

\subsection{A look to the future}

Interestingly, Gaussian process regression --- essentially the same method as kriging under a different name --- is today considered a leading machine learning technique for applications where uncertainty quantification is important, and is often applied to higher dimensional problems than the mostly-spatial ones encountered in the field of geostatistics. Gaussian process regression is favoured over other methods for uncertainty quantification due to its well understood mathematical properties and its compatibility with the Bayesian framework \cite{gibbs1998bayesian} (although geostatistics has tended to remain frequentist - with the BLUP being a frequentist concept) . However, in 1995 Radford Neal showed that as the number of hidden nodes in a single layer fully-connected neural network approaches infinity, the network will become mathematically equivalent to a Gaussian process \cite{neal1995bayesian}. More recently, similar equivalence has been explored between deep fully-connected neural networks and Gaussian processes \cite{lee2017deep}, and we now have Deep Gaussian processes \cite{damianou2013deep}, including with convolutional layers \cite{blomqvist2019deep}. So what's the catch? Computational complexity. In terms of time, neural network training scales linearly with the number of observations, however Gaussian process inference scales with the cube of the number of observations \cite{liu2020gaussian}. This is perhaps the main reason why deep neural networks have risen into mainstream applications ahead of Gaussian process regression - they are allowing us to solve otherwise unsolvable big data problems, and in many applications deterministic prediction is adequate. However, as the neural network community strives to improve their ability to quantify uncertainty, and the Gaussian process community strives to reduce their computational footprint, the two camps may well converge on methods that provide very similar functionality to practitioners.

Where will this leave geostatistics? It seems important to frame the functionality of well established geostatistical models (\autoref{kriging}) in the context of the functionality that deep neural networks (and deep Gaussian processes) can bring to the table. As demonstrated in this paper, the capability to optimally extract information from unstructured data (like terrain grids) is extremely powerful, and could be a game changer in terms of eliminating the need to manually design (sub-optimal) covariates for use in geostatistical analyses. Essentially, we can push our variable selection processes right back to whatever raw unstructured data we have available, and trust (with empirical evaluation) deep neural networks to extract the relevant information. There is an argument to say that doing so reduces the interpretability of our model (deep learning as a `black box'), which we may have wished to preserve. It is true that there is no way to effectively convey or comprehend in `explainable' terms the series of transformations that our neural network applies to terrain texture in order to produce a representation that correlates maximally with the target variable. On the other hand, if deep learning allows us to explain a higher proportion of variance without deferring to spatial auto-correlation, which is itself fairly opaque, then that could be seen as beneficial. With models increasingly being used to support important decision making, it could be argued that models should be judged by the quality of information they provide, rather than by how easily interpretable they are, in order that we can progress towards optimal decision making. In the end, the best approach to choose will be the one that best satisfies the objectives at hand, and this will always be case dependent.

\section{Conclusions}

In this paper we have demonstrated a new approach for utilising deep learning to derive optimal terrain texture covariates for geostatistical modelling applications. We have shown that our deep learning approach is entirely compatible with the typical geostatistical model formulation (\autoref{kriging}) and in fact can be used as the exclusive source of covariate information in a `deep covariate-learning' geostatistical model formulation (\autoref{dcl}). The results our deep neural network achieves on held-out test data are extremely encouraging. Terrain data has historically not tended to be regarded as particularly informative for most geochemical applications, at least within quantitative modelling, and yet our deep neural network has been able to extract sufficient information to explain 61\%, 58\% and 64\% of the variance of our target variables: log(arsenic), calcium. and potassium concentrations in stream sediments. This is all from using \textit{only} terrain texture, without accounting for spatial variability explicitly (the network was not provided with easting, northing, and absolute elevation as inputs, and had only 16x16km square images of terrain texture to work with). Within the geostatistical modelling framework, this spatial variability is accounted for instead by the spatial random variable component of the model ($u$ in \autoref{kriging}).

Our results suggest that deep learning has a very significant role to play in the future of geostatistical modelling, and offers a step-change in how we can make use of gridded auxiliary datasets in the modelling process by allowing us to optimise the extraction of information from them. The covariates that our deep learning approach learns to derive are spatially generalisable within the study area, and it is quite possible that they can shed new predictive light on otherwise under-sampled geographic regions, for example for mineral exploration purposes. The strong predictive performance achieved using only 16x16km windows of terrain texture warrants further investigation of our deep covariate-learning approach using different window sizes, as well as including channels for additional auxilliary variables. The apparent ability of deep learning to capture complex structural relationships (for example, appearing to realise that stream sediments do `flow' from upstream catchments) mean it will be interesting to see how much further this approach can be developed. If enough explanation can be obtained from gridded datasets alone, then perhaps we will no longer have need for the spatial random variable component of our geostatistical models ($u$ in \autoref{kriging}), and the fairly opaque spatial autocorrelation based explanation that it provides. Alternatively, we may find that the best overall predictive performance is achieved by using deep learning end-to-end for geostatistical modelling tasks, in which it would have the benefit over the typical geostatistical model (\autoref{kriging}) of being able to learn interactions between covariates features (which are themselves learned) and spatial location. Reliable estimates of uncertainty are perhaps the main justification for refraining from an end-to-end deep learning approach at the moment, hence in this paper we demonstrate deep learning within the typical geostatistical modelling framework, but research into improving uncertainty quantification in deep learning is developing at a rapid pace. If the future of transport will be dominated by autonomous vehicles, the future of geostatistical modelling will surely also be driven by deep learning.

\subsection*{Acknowledgements}
We acknowledge funding from the UK's Engineering and Physical Sciences Research Council (EPSRC project ref: 2071900). Our thanks go to Nvidia and their GPU grant scheme for kindly providing us with a Titan X Pascal GPU to accelerate our neural network training in this study.

We also thank the British Geological Survey for making the G-BASE geochemical data available for this study. For academic research purposes, readers may request access to the G-BASE dataset from the British Geological Survey at https://www.bgs.ac.uk/enquiries/home.html or by email to enquiries@bgs.ac.uk.

\subsection*{Declaration of interests}
The authors declare that they have no known competing financial interests or personal relationships that could have appeared to influence the work reported in this paper.

\bibliographystyle{unsrtnat}  
\bibliography{references}  

\begin{thebibliography}{42}
\providecommand{\natexlab}[1]{#1}
\providecommand{\url}[1]{\texttt{#1}}
\expandafter\ifx\csname urlstyle\endcsname\relax
  \providecommand{\doi}[1]{doi: #1}\else
  \providecommand{\doi}{doi: \begingroup \urlstyle{rm}\Url}\fi

\bibitem[Krige(1951)]{krige1951statistical}
Daniel~G Krige.
\newblock A statistical approach to some basic mine valuation problems on the
  witwatersrand.
\newblock \emph{Journal of the Southern African Institute of Mining and
  Metallurgy}, 52\penalty0 (6):\penalty0 119--139, 1951.

\bibitem[Tobler(1970)]{tobler1970computer}
Waldo~R Tobler.
\newblock A computer movie simulating urban growth in the detroit region.
\newblock \emph{Economic geography}, 46\penalty0 (sup1):\penalty0 234--240,
  1970.

\bibitem[Matheron(1969)]{matheron1969krigeage}
G~Matheron.
\newblock \emph{Le krigeage universel: cahiers du Centre de Morphologie
  Mathematique}.
\newblock École nationale supérieure des mines de Paris, 1969.

\bibitem[Odeh et~al.(1995)Odeh, McBratney, and Chittleborough]{odeh1995further}
Inakwu~OA Odeh, AB~McBratney, and DJ~Chittleborough.
\newblock Further results on prediction of soil properties from terrain
  attributes: heterotopic cokriging and regression-kriging.
\newblock \emph{Geoderma}, 67\penalty0 (3-4):\penalty0 215--226, 1995.

\bibitem[Hudson and Wackernagel(1994)]{hudson1994mapping}
Gordon Hudson and Hans Wackernagel.
\newblock Mapping temperature using kriging with external drift: theory and an
  example from scotland.
\newblock \emph{International journal of Climatology}, 14\penalty0
  (1):\penalty0 77--91, 1994.

\bibitem[Lark(2012)]{lark2012towards}
RM~Lark.
\newblock Towards soil geostatistics.
\newblock \emph{Spatial Statistics}, 1:\penalty0 92--99, 2012.

\bibitem[Stein(1999)]{stein1999interpolation}
Michael~L Stein.
\newblock \emph{Interpolation of spatial data: some theory for kriging}.
\newblock Springer, New York, 1999.

\bibitem[Lark(2000)]{lark2000estimating}
RM~Lark.
\newblock Estimating variograms of soil properties by the method-of-moments and
  maximum likelihood.
\newblock \emph{European Journal of Soil Science}, 51\penalty0 (4):\penalty0
  717--728, 2000.

\bibitem[Riley et~al.(1999)Riley, DeGloria, and Elliot]{riley1999index}
Shawn~J Riley, Stephen~D DeGloria, and Robert Elliot.
\newblock Index that quantifies topographic heterogeneity.
\newblock \emph{intermountain Journal of sciences}, 5\penalty0 (1-4):\penalty0
  23--27, 1999.

\bibitem[Berti et~al.(2013)Berti, Corsini, and Daehne]{berti2013comparative}
Matteo Berti, Alessandro Corsini, and Alexander Daehne.
\newblock Comparative analysis of surface roughness algorithms for the
  identification of active landslides.
\newblock \emph{Geomorphology}, 182:\penalty0 1--18, 2013.

\bibitem[Stambaugh and Guyette(2008)]{stambaugh2008predicting}
Michael~C Stambaugh and Richard~P Guyette.
\newblock Predicting spatio-temporal variability in fire return intervals using
  a topographic roughness index.
\newblock \emph{Forest Ecology and Management}, 254\penalty0 (3):\penalty0
  463--473, 2008.

\bibitem[Milodowski et~al.(2015)Milodowski, Mudd, and
  Mitchard]{milodowski2015topographic}
DT~Milodowski, SM~Mudd, and ETA Mitchard.
\newblock Topographic roughness as a signature of the emergence of bedrock in
  eroding landscapes.
\newblock \emph{Earth Surface Dynamics}, 3\penalty0 (4):\penalty0 483--499,
  2015.

\bibitem[Krizhevsky et~al.(2012)Krizhevsky, Sutskever, and
  Hinton]{krizhevsky2012imagenet}
Alex Krizhevsky, Ilya Sutskever, and Geoffrey~E Hinton.
\newblock Imagenet classification with deep convolutional neural networks.
\newblock In \emph{Advances in neural information processing systems}, pages
  1097--1105, 2012.

\bibitem[Han et~al.(2014)Han, Zhang, Cheng, Guo, and Ren]{han2014object}
Junwei Han, Dingwen Zhang, Gong Cheng, Lei Guo, and Jinchang Ren.
\newblock Object detection in optical remote sensing images based on weakly
  supervised learning and high-level feature learning.
\newblock \emph{IEEE Transactions on Geoscience and Remote Sensing},
  53\penalty0 (6):\penalty0 3325--3337, 2014.

\bibitem[Zou et~al.(2015)Zou, Ni, Zhang, and Wang]{zou2015deep}
Qin Zou, Lihao Ni, Tong Zhang, and Qian Wang.
\newblock Deep learning based feature selection for remote sensing scene
  classification.
\newblock \emph{IEEE Geoscience and Remote Sensing Letters}, 12\penalty0
  (11):\penalty0 2321--2325, 2015.

\bibitem[Zhang et~al.(2016)Zhang, Zhang, and Du]{zhang2016deep}
Liangpei Zhang, Lefei Zhang, and Bo~Du.
\newblock Deep learning for remote sensing data: A technical tutorial on the
  state of the art.
\newblock \emph{IEEE Geoscience and Remote Sensing Magazine}, 4\penalty0
  (2):\penalty0 22--40, 2016.

\bibitem[Zhu et~al.(2017)Zhu, Tuia, Mou, Xia, Zhang, Xu, and
  Fraundorfer]{zhu2017deep}
Xiao~Xiang Zhu, Devis Tuia, Lichao Mou, Gui-Song Xia, Liangpei Zhang, Feng Xu,
  and Friedrich Fraundorfer.
\newblock Deep learning in remote sensing: A comprehensive review and list of
  resources.
\newblock \emph{IEEE Geoscience and Remote Sensing Magazine}, 5\penalty0
  (4):\penalty0 8--36, 2017.

\bibitem[Ma et~al.(2019)Ma, Liu, Zhang, Ye, Yin, and Johnson]{ma2019deep}
Lei Ma, Yu~Liu, Xueliang Zhang, Yuanxin Ye, Gaofei Yin, and Brian~Alan Johnson.
\newblock Deep learning in remote sensing applications: A meta-analysis and
  review.
\newblock \emph{ISPRS journal of photogrammetry and remote sensing},
  152:\penalty0 166--177, 2019.

\bibitem[Li et~al.(2011)Li, Heap, Potter, and Daniell]{li2011application}
Jin Li, Andrew~D Heap, Anna Potter, and James~J Daniell.
\newblock Application of machine learning methods to spatial interpolation of
  environmental variables.
\newblock \emph{Environmental Modelling \& Software}, 26\penalty0
  (12):\penalty0 1647--1659, 2011.

\bibitem[Pourghasemi and Rahmati(2018)]{pourghasemi2018prediction}
Hamid~Reza Pourghasemi and Omid Rahmati.
\newblock Prediction of the landslide susceptibility: Which algorithm, which
  precision?
\newblock \emph{Catena}, 162:\penalty0 177--192, 2018.

\bibitem[Rodriguez-Galiano et~al.(2015)Rodriguez-Galiano, Sanchez-Castillo,
  Chica-Olmo, and Chica-Rivas]{rodriguez2015machine}
V~Rodriguez-Galiano, M~Sanchez-Castillo, M~Chica-Olmo, and MJOGR Chica-Rivas.
\newblock Machine learning predictive models for mineral prospectivity: An
  evaluation of neural networks, random forest, regression trees and support
  vector machines.
\newblock \emph{Ore Geology Reviews}, 71:\penalty0 804--818, 2015.

\bibitem[Wadoux(2019)]{wadoux2019using}
Alexandre MJ-C Wadoux.
\newblock Using deep learning for multivariate mapping of soil with quantified
  uncertainty.
\newblock \emph{Geoderma}, 351:\penalty0 59--70, 2019.

\bibitem[Wadoux et~al.(2019)Wadoux, Padarian, and Minasny]{wadoux2019multi}
Alexandre M~JC Wadoux, Jos{\'e} Padarian, and Budiman Minasny.
\newblock Multi-source data integration for soil mapping using deep learning.
\newblock \emph{Soil}, 5\penalty0 (1):\penalty0 107--119, 2019.

\bibitem[Padarian et~al.(2019)Padarian, Minasny, and https://orcidorg/0000
  0003-0913-2643]{padarian2019using}
José Padarian, Budiman Minasny, and Alex B~McBratney https://orcidorg/0000
  0003-0913-2643.
\newblock Using deep learning for digital soil mapping.
\newblock \emph{Soil}, 5\penalty0 (1):\penalty0 79--89, 2019.

\bibitem[Van~Zyl(2001)]{van2001shuttle}
Jakob~J Van~Zyl.
\newblock The shuttle radar topography mission (srtm): a breakthrough in remote
  sensing of topography.
\newblock \emph{Acta Astronautica}, 48\penalty0 (5-12):\penalty0 559--565,
  2001.

\bibitem[Johnson et~al.(2005)Johnson, Breward, Ander, and Ault]{johnson2005g}
CC~Johnson, N~Breward, EL~Ander, and L~Ault.
\newblock G-base: baseline geochemical mapping of great britain and northern
  ireland.
\newblock \emph{Geochemistry: exploration, environment, analysis}, 5\penalty0
  (4):\penalty0 347--357, 2005.

\bibitem[Kirkwood et~al.(2016{\natexlab{a}})Kirkwood, Everett, Ferreira, and
  Lister]{kirkwood2016stream}
Charlie Kirkwood, Paul Everett, Antonio Ferreira, and Bob Lister.
\newblock Stream sediment geochemistry as a tool for enhancing geological
  understanding: An overview of new data from south west england.
\newblock \emph{Journal of Geochemical Exploration}, 163:\penalty0 28--40,
  2016{\natexlab{a}}.

\bibitem[Pawlowsky-Glahn et~al.(2015)Pawlowsky-Glahn, Egozcue, and
  Tolosana-Delgado]{pawlowsky2015modeling}
Vera Pawlowsky-Glahn, Juan~Jos{\'e} Egozcue, and Raimon Tolosana-Delgado.
\newblock \emph{Modeling and analysis of compositional data}.
\newblock John Wiley \& Sons, 2015.

\bibitem[Kim et~al.(2017)Kim, Choi, Yi, and Park]{kim2017geostatistical}
Sung-Min Kim, Yosoon Choi, Huiuk Yi, and Hyeong-Dong Park.
\newblock Geostatistical prediction of heavy metal concentrations in stream
  sediments considering the stream networks.
\newblock \emph{Environmental Earth Sciences}, 76\penalty0 (2):\penalty0 72,
  2017.

\bibitem[Kirkwood(2016)]{kirkwood2016dropout}
Charlie Kirkwood.
\newblock {A dropout-regularised neural network for mapping arsenic enrichment
  in SW England using MXNet}, 2016.

\bibitem[Kirkwood et~al.(2016{\natexlab{b}})Kirkwood, Cave, Beamish, Grebby,
  and Ferreira]{kirkwood2016machine}
Charlie Kirkwood, Mark Cave, David Beamish, Stephen Grebby, and Antonio
  Ferreira.
\newblock A machine learning approach to geochemical mapping.
\newblock \emph{Journal of Geochemical Exploration}, 167:\penalty0 49--61,
  2016{\natexlab{b}}.

\bibitem[Zuo(2017)]{zuo2017machine}
Renguang Zuo.
\newblock Machine learning of mineralization-related geochemical anomalies: A
  review of potential methods.
\newblock \emph{Natural Resources Research}, 26\penalty0 (4):\penalty0
  457--464, 2017.

\bibitem[Kirkwood et~al.(2017)Kirkwood, Cooper, Ferreria, and
  Beamish]{kirkwood2017unmixing}
Charlie Kirkwood, Mark Cooper, Antonio Ferreria, and David Beamish.
\newblock Unmixing and mapping components of northern ireland’s geochemical
  composition using fastica and random forests.
\newblock \emph{EarthArXiv preprint http://eartharxiv.org/8k3f7/}, 2017.

\bibitem[Gal and Ghahramani(2015)]{gal2015dropout}
Yarin Gal and Zoubin Ghahramani.
\newblock Dropout as a bayesian approximation: Representing model uncertainty
  in deep learning.
\newblock \emph{arXiv preprint arXiv:1506.02142}, 2015.

\bibitem[Kendall and Gal(2017)]{kendall2017uncertainties}
Alex Kendall and Yarin Gal.
\newblock What uncertainties do we need in bayesian deep learning for computer
  vision?
\newblock In \emph{Advances in neural information processing systems}, pages
  5574--5584, 2017.

\bibitem[Farquhar et~al.(2019)Farquhar, Osborne, and Gal]{farquhar2019radial}
Sebastian Farquhar, Michael Osborne, and Yarin Gal.
\newblock Radial bayesian neural networks: Beyond discrete support in
  large-scale bayesian deep learning, 2019.

\bibitem[Gibbs(1998)]{gibbs1998bayesian}
Mark~N Gibbs.
\newblock \emph{Bayesian Gaussian processes for regression and classification}.
\newblock PhD thesis, Citeseer, 1998.

\bibitem[Neal(1995)]{neal1995bayesian}
Radford~M Neal.
\newblock \emph{Bayesian Learning For Neural Networks}.
\newblock PhD thesis, University of Toronto, 1995.

\bibitem[Lee et~al.(2017)Lee, Bahri, Novak, Schoenholz, Pennington, and
  Sohl-Dickstein]{lee2017deep}
Jaehoon Lee, Yasaman Bahri, Roman Novak, Samuel~S Schoenholz, Jeffrey
  Pennington, and Jascha Sohl-Dickstein.
\newblock Deep neural networks as gaussian processes.
\newblock \emph{arXiv preprint arXiv:1711.00165}, 2017.

\bibitem[Damianou and Lawrence(2013)]{damianou2013deep}
Andreas Damianou and Neil Lawrence.
\newblock Deep gaussian processes.
\newblock In \emph{Artificial Intelligence and Statistics}, pages 207--215,
  2013.

\bibitem[Blomqvist et~al.(2019)Blomqvist, Kaski, and
  Heinonen]{blomqvist2019deep}
Kenneth Blomqvist, Samuel Kaski, and Markus Heinonen.
\newblock Deep convolutional gaussian processes.
\newblock In \emph{Joint European Conference on Machine Learning and Knowledge
  Discovery in Databases}, pages 582--597. Springer, 2019.

\bibitem[Liu et~al.(2020)Liu, Ong, Shen, and Cai]{liu2020gaussian}
Haitao Liu, Yew-Soon Ong, Xiaobo Shen, and Jianfei Cai.
\newblock When gaussian process meets big data: A review of scalable gps.
\newblock \emph{IEEE Transactions on Neural Networks and Learning Systems},
  2020.

\end{thebibliography}



\end{document}